\documentclass[10pt,twoside]{article}

\newlength\titlebox 
\skip\footins 9pt plus 4pt minus 2pt
\def\footnoterule{\kern-3pt \hrule width 5pc \kern 2.6pt }
\labelwidth\leftmargini\advance\labelwidth-\labelsep \labelsep 5pt

\usepackage{sectsty}
\sectionfont{\fontsize{12}{15}\selectfont}
\subsectionfont{\fontsize{10}{15}\selectfont}

\usepackage[numbers,sectionbib]{natbib}

\usepackage{wrapfig}

\usepackage{hyperref}
\usepackage{url}

\usepackage{graphicx}
\usepackage{color}
\usepackage{rotating}
\usepackage{bbm}
\usepackage{latexsym}
\usepackage{amsmath}               
\usepackage{amssymb}               
\usepackage{psfrag} 
\usepackage{wasysym}
\usepackage{epstopdf}

\usepackage[ruled,vlined]{algorithm2e}

\definecolor{brightred}{rgb}{1.0,0.1,0.1}
\definecolor{brightblue}{rgb}{0.0,0.0,0.8}
\definecolor{darkblue}{rgb}{0.0,0.0,0.5}
\definecolor{darkgreen}{rgb}{0.0,0.3,0.0}
\definecolor{brightgreen}{rgb}{0.0,0.8,0.0}
\definecolor{darkblack}{rgb}{0.0,0.0,0.0}
\definecolor{grey}{rgb}{0.3,0.3,0.3}

\graphicspath{{./}}
\DeclareGraphicsExtensions{.eps,.png}
\newcommand{\dblone}{\hbox{$1\hskip -1.2pt\vrule depth 0pt height 1.6ex width 0.7pt\vrule depth 0pt height 0.3pt width 0.12em$}}

\newcommand{\FF}{{\cal F}}

\newcommand{\disS}{\displaystyle}

\newcommand{\refp}[1]{(\ref{#1})}
\newcommand{\sVec}{\vec{s}}

\newcommand{\sVecNew}{\vec{s}_{\mathrm{new}}}
\newcommand{\sVecPrime}{\vec{s}^{\,\prime}}
\newcommand{\sVecT}{\sVecPrime}

\newcommand{\yVec}{\vec{y}}
\newcommand{\yVecN}{\vec{y}^{\,(n)}}

\newcommand{\E}[1]{\left\langle{}#1\right\rangle}

\newcommand{\KK}{{\cal K}}
\newcommand{\KKn}{\KK^{\hspace{-0.8ex}\phantom{1}^{(n)}}}
\newcommand{\KKnew}{\KK_{\mathrm{new}}}

\newcommand{\KKnnew}{\KKn_{\mathrm{new}}}

\newcommand{\comment}[1]{}

\newcommand{\qn}{q^{(n)}}

\newcommand{\beq}{\begin{equation}}
\newcommand{\eeq}{\end{equation}}
\newcommand{\beqo}{\begin{displaymath}}
\newcommand{\eeqo}{\end{displaymath}}
\newcommand{\bea}{\begin{eqnarray}}
\newcommand{\eea}{\end{eqnarray}} 
\newcommand{\beao}{\begin{eqnarray*}}
\newcommand{\eeao}{\end{eqnarray*}}

\newcommand{\One}{\dblone}

\newcommand{\pVarN}{p^{(n)}_{\mathrm{var}}}

\newcommand{\qMar}{q_{\mathrm{mar}}}
\newcommand{\qMarN}{q^{(n)}_{\mathrm{mar}}}

\newcommand{\calO}{{\cal O}}

\newcommand{\myvanish}[1]{}

\begin{document}
%
%
%
%
\newcommand{\disttt}{\hspace{9em}}
\newcommand{\mmm}{\vspace{3mm}}
\title{Truncated Variational Sampling for\\
`Black Box' Optimization of Generative Models\\ \ }
%
%
%
\author{J\"org L\"ucke \\
{\tt joerg.luecke@uol.de}
              \\
                     Universit\"at Oldenburg and\\
                     Cluster of Excellence H4a\\
                     Oldenburg, Germany\\
\and Zhenwen Dai \\
{\tt zhenwend@amazon.com}\\
Amazon Research\\
Cambridge, UK
\and Georgios Exarchakis\\
{\tt georgios.exarchakis@ens.fr}\\
Department d'Informatique\\
\'Ecole Normale Sup\'erieure\\
Paris, France
}
%
%

%
%
%
\date{\phantom{nothing}}
\maketitle
\begin{abstract}%
\noindent
We investigate the optimization of two probabilistic generative models with binary latent variables using a novel variational EM approach. The approach distinguishes itself from previous variational approaches by using latent states as variational parameters. Here we use efficient and general purpose sampling procedures to vary the latent states, and investigate the `black box' applicability of the resulting optimization procedure. For general purpose applicability, samples are drawn from approximate marginal distributions of the considered generative model as well as from the model's prior distribution. As such, variational sampling is defined in a generic form, and is directly executable for a given model. As a proof of concept, we then apply the novel procedure (A) to Binary Sparse Coding (a model with continuous observables), and (B) to basic Sigmoid Belief Networks (which are models with binary observables). Numerical experiments verify that the investigated approach efficiently as well as effectively increases a variational free energy objective without requiring any additional analytical steps.\vspace{2mm}
\end{abstract}

\section{Introduction\vspace{0mm}}
The use of expectation maximization (EM) for advanced probabilistic data models requires approximations
because EM with an exact E-step (computing the full posterior) is typically intractable. Many models of recent interest have binary latents \citep[][]{GoodfellowEtAl2012,GanEtAl2015,MnihGregor2014,SheikhLucke2016}, 
and for such models these intractabilities are primarily computational: exact E-steps can be computed but they scale exponentially with the number of latents.
To overcome intractabilities, overcome intractabilities for models with binary latents there are typically
two types of approaches applied: sampling approaches or variational EM with the latter having been dominated by factored variational
approaches in the past \citep[e.g.][]{JordanEtAl1999}. Variational approaches and sampling have also often been combined
\citep[][]{HoffmanBlei2015,SalimansEtAl2015,HernandezEtAl2016,SheikhLucke2016} to leverage
the advantages of both methods. However, given a generative model, both approximations require often cumbersome
derivations either to derive efficient posterior samplers or to derive update equations for variational parameter optimization. 
The question how procedures can be defined that automatize the development of learning algorithms for generative models has therefore
shifted into the focus of recent research \citep[][]{RanganathEtAl2014,TranEtAl2015,RezendeMohamed2015,KucukelbirEtAl2016,SheikhLucke2016}.
In this paper, we make use of truncated approximations to EM which have repeatedly been applied before \citep[][]{LuckeEggert2010,SheikhLucke2016,ForsterLucke2018}.
Here we show how novel theoretical results on truncated variational distributions \citep[][]{Lucke2016} can be used to couple variational EM and sampling exceptionally tightly.
This coupling then enables ``black box'' applicability.\vspace{-1mm}
\section{Truncated Posteriors and Sampling\vspace{-2mm}}
Let us consider generative models with $H$ binary latent variables, $\sVec=(s_1,\ldots,s_H)$ with $s_h\in\{0,1\}$.
%
%
Truncated approximations have been motivated by the observation
that the exponentially large sums over states required for expectation values w.r.t.\,posteriors are typically dominated by summands
corresponding to very few states. If for a given data point $\yVecN$ these few states are contained in a set $\KKn$, we
can define a posterior approximation as follows \citep[compare][]{LuckeEggert2010,SheikhLucke2016}:
 \begin{eqnarray}
\label{EqnQMain}
 \qn(\sVec;\KK,\Theta) = \frac{\disS\phantom{\int}p(\sVec\,|\,\yVecN,\Theta)\,\delta(\sVec\in\KKn)\phantom{\int}}
 {\disS\sum_{\sVecT\in\KKn}p(\sVecT\,|\,\yVecN,\Theta)}\,,
 \end{eqnarray}
where $\delta(\sVec\in\KKn)=1$ if $\KKn$ contains $\sVec$ and zero otherwise.
It is straight-forward to derive expectation values w.r.t.\,these approximate posteriors simply by inserting \refp{EqnQMain}
into the definition of expectation values and by multiplying numerator and denominator by $p(\yVecN\,|\,\Theta)$, which yields:
\begin{equation}
\E{g(\sVec)}_{\qn}
%
%
\,=\,\frac{\disS\sum_{\sVec\in\KKn} p(\sVec,\yVecN\,|\,\Theta)\ g(\sVec)
}{\disS\sum_{\sVecT\in\KKn}p(\sVecT,\yVecN\,|\,\Theta)}
\label{EqnSuffStat}
\end{equation}
where $g(\sVec)$ is a function of the hidden variables. As the dominating summands are different
for each data point $\yVecN$, the sets $\KKn$ are different. If a set $\KKn$ now contains those
states $\sVec$ which dominate the sums over the joints w.r.t.\ the exact posterior, then
Eqn.\,\ref{EqnSuffStat} is a very accurate approximation. 

Truncated posterior approximations have successfully been applied to a number of elementary and more advanced
generative models, and they do not suffer from potential biases
introduced by posterior independence assumptions made by factored variational approximations 
Previously, the sets $\KKn$ were defined based on sparsity assumptions and/or latent
preselection \citep[][]{LuckeEggert2010,SheltonEtAl2017}.
The approach followed here, in contrast, uses sets $\KKn$ which contain samples from model and data dependent distributions. By treating
the truncated distribution (\ref{EqnQMain}) as variational distributions within a free-energy framework \citep[][]{Lucke2016}, we can then
derive efficient procedures to update the samples in $\KKn$ such that the variational free-energy is always monotonically increased.
For this we use the following theoretical results: (1)~We use that the M-step equations
remain unchanged if instead
of exact posteriors the truncated posteriors (\ref{EqnQMain}) are used; (2)~We make use of the result that after each M-step the free-energy
corresponding to truncated variational distributions is given by the following simplified and computationally tractable form:\vspace{4mm}
%
 \begin{equation}
\disS\FF(\KK,\Theta)\,=\,\sum_{n}\ \log\big(\sum_{\sVec\in\KKn}\ p(\sVec,\yVecN\,|\,\Theta)\ \big)\,,\vspace{3mm}
\label{EqnFreeEnergy}
 \end{equation}
where $\KK=(\KK^{(1)},\ldots,\KK^{(N)})$. The variational E-step then consists of finding a set $\KKnew$ which increases $\FF(\KK,\Theta)$
w.r.t.\ $\KK$. The M-step consist of the standard M-step equations but with expectation values estimated by (\ref{EqnSuffStat}). 

For any larger scale multiple-cause model we can not exhaustively iterate through all latent states. We therefore here seek to find new sets $\tilde{\KK}$ using sampling, such that the free-energy is increased, $\FF(\tilde{\KK},\Theta) > \FF(\KK,\Theta)$. To keep the computational demand limited, we will take the sets $\KK$ and $\tilde{\KK}$ to be of
constant size after each E-step by demanding $|\KKn|=|\tilde{\KK}^{(n)}|=S$ for all $n$. Instead of explicitly computing and comparing the free-energies (\ref{EqnFreeEnergy})
w.r.t.\,$\KK$ and $\tilde{\KK}$, we can instead use a comparison of joint probabilities $p(\sVec,\,\yVecN\,|\,\Theta)$ as a criterion for
free-energy increase. The following can be shown \citep[][]{Lucke2016}: 

For a replacement of $\sVec\in\KKn$ by a new state \mbox{$\sVecNew\not\in\KKn$} the free-energy $\FF(\KK,\Theta)$ is increased if and only if\vspace{4mm}
\begin{equation}
%
p(\sVecNew,\,\yVecN\,|\,\Theta) > p(\sVec,\,\yVecN\,|\,\Theta)\,.\vspace{4mm}
\label{EqnCriterion}
\end{equation}
Criterion (\ref{EqnCriterion}) may directly be concluded by considering the functional form of Eqn.\,\ref{EqnFreeEnergy} \citep[see][for a formal proof]{Lucke2016}. It means that the free-energy is guaranteed to increase if we replace, e.g., the state with the lowest joint in $\KKn$
by a newly sampled state $\sVecNew\not\in\KKn$ with a higher joint. Instead of comparing single joints, a computationally more efficient procedure
is to use batches of many newly sampled states, and then to use criterion (\ref{EqnCriterion}) to increase
\newcommand{\algBreak}{\vspace{0mm}\\}
\begin{wrapfigure}[10]{r}{0.65\textwidth}
\begin{minipage}{0.64\textwidth}
\ \\[-1.5mm]
\SetAlCapHSkip{0.2em}
\DecMargin{-1.0em}
\begin{algorithm}[H]\vspace{1.5mm}
%
%
\For{$n=1,\ldots,N$\vspace{1mm}}{
draw $M$ samples $\sVec\,\sim\,\pVarN(\sVec)$;\algBreak
define $\KKnnew$ to contain all $M$ samples;\algBreak
set $\KKn=\KKn \cup \KKnnew$;\algBreak
remove those $(|\KKn|-S)$ samples $\sVec\in\KKn$ 
with the lowest $p(\sVec,\,\yVecN\,|\,\Theta)$;\vspace{1mm}
}
%
%
\caption{Sampling-based TV-E-step\label{AlgEStep}}
\end{algorithm}
\end{minipage}
\end{wrapfigure}
 $\FF(\KK,\Theta)$ as much as possible. Such a procedure is given by Alg.\,\ref{AlgEStep}: For each data point $n$, we first draw $M$ new samples from a yet to be specified distribution $\pVarN(\sVec)$.
These samples are then united with the states already in $\KKn$. Of this union of old and new states, we then take the $S$ states with highest joints
to define the new state set $\KKn$. This last step selects because of (\ref{EqnCriterion}) the best possible subset of the union. Furthermore, selecting
the $S$ states with largest joints represents a standard unsorted partial sorting problem which is solvable in linear time complexity, i.e., with at most
${\calO}(M+S)$ in our case. Instead of selecting the $S$ largest joints, we can also remove the $(|\KKn|-S)$ lowest ones (last line
in Alg.\,\ref{AlgEStep}). For {\em any} distribution $\pVarN(\sVec)$, Alg.\,\ref{AlgEStep} is guaranteed to monotonously increase the free-energy
$\FF(\KK,\Theta)$ w.r.t.\ $\KK$.\vspace{-2mm}
%
%
%
%
%
%
%

\section{\hspace{0mm}Posterior, Prior, and Marginal Sampling\vspace{-2mm}} \label{sec:marginal_sampling}

While the partial E-step of Alg.\,\ref{AlgEStep} monotonously increases the free-energy for any distribution $\pVarN(\sVec)$ used for sampling,
the specific choice for $\pVarN(\sVec)$ is of central importance for the efficiency of the procedure. If the distribution is not chosen well, 
any significant increase of $\FF(\KK,\Theta)$ may require unreasonable amounts of time, e.g., because new samples which increase $\FF(\KK,\Theta)$
are sampled too infrequently. By considering Alg.\,\ref{AlgEStep}, the requirement for $\pVarN(\Theta)$ is to provide samples with high joint probability 
$p(\sVec,\,\yVecN\,|\,\Theta)$ for a given $\yVecN$. The first distribution that comes to 
mind for $\pVarN(\Theta)$ is the posterior distribution $p(\sVec\,|\,\yVecN,\Theta)$. Samples
form the posterior are likely to have high posterior mass and therefore high joint mass relative to the other states because
all states share the same normalizer $p(\yVecN\,|\,\Theta)$.
On the downside, however, sampling from the posterior may not be an easy task for models with binary latents and a relatively
high dimensionality as we intend to aim at here. Furthermore, the derivation of posterior samplers requires additional analytical
efforts for any new generative model we apply the procedure to, and requires potentially additional design choices such as
definitions of proposal distributions. All these points are contrary to our goal of a `black box' procedure which is applicable as generally and
generically as possible. Instead, we therefore seek distributions $\pVarN(\Theta)$ for Alg.\,\ref{AlgEStep} that can
efficiently optimize the free-energy but that can be defined without requiring model-specific analytical derivations.
Candidates for $\pVarN(\Theta)$ are consequently the prior distribution of
the given generative model, $p(\sVec\,|\,\Theta)$, or the marginal distribution. 
A prior sampler is usually directly given by the generative model but may have the disadvantage that finally new
samples only very rarely increase the free-energy because the prior sampler is independent of a given data point (only the average
over data points has high posterior mass). Marginal samplers, on the other hand, are data driven but the computation of
activation probabilities $p(s_h=1\,|\,\yVecN,\Theta)$ is unfortunately not computationally efficient.
To obtain data-driven but efficient samplers, we will for our purposes, therefore, use approximate marginal samplers. 
%

{\bf 1st Approximation.} First observe that we can obtain an efficiently computable approximation to a marginal
sampler by using the truncated distributions $\qn(\sVec)$ in \refp{EqnQMain} themselves. For binary latents $s_h$ we can approximate:
\begin{equation}
p(s_h=1\,|\,\yVecN,\Theta) = \E{s_h}_{p(\sVec\,|\,\yVecN,\Theta)}
			\approx \E{s_h}_{\qn(\sVec)}, 
\label{EqnMarginalOne}
\end{equation}
and accordingly $p(s_h=0\,|\,\yVecN,\Theta)$.
%
%
%
Because of the arguments given above the expectations (\ref{EqnSuffStat}) w.r.t.\ $\qn(\sVec)$ are efficiently computable using (\ref{EqnSuffStat}) with $g(\sVec)=\sVec$.
Using \ref{EqnMarginalOne} we can consequently define for each latent $h$ an approximation of the marginal $p(s_h=1\,|\,\yVecN,\Theta)$.
Given a directed generative model, no derivations are required to efficiently generate samples from this approximation
because the joint probabilities to estimate $p(s_h=1\,|\,\yVecN,\Theta)$ using (\ref{EqnSuffStat}) can 
directly be computed. The truncated marginal sampler defined by Eqn.\,\ref{EqnMarginalOne} becomes increasingly similar to an exact marginal sampler
the better the truncated distributions approximate the exact posteriors.

{\bf 2nd Approximation.} To further improve efficiency and convergence times, we optionally apply a second approximation by using the approximate
marginal distributions (Eqns.\,\ref{EqnMarginalOne}) as target objective for a parametric function
$f_h(\yVecN;\Lambda)$ which approximates the truncated marginal. A parametric function from data to marginal probabilities of the latents has the advantage
of modeling data similarities by mapping similar data to similar marginal distributions. The mapping incorporates information
across the data points, which can facilitate training and, e.g., avoids more expensive $\KKn$ updates of some data points due to bad initialization.
The mapping $f_h(\yVecN;\Lambda)$ is estimated with the training data and the current approximate marginal $\qMar(s_h=1\,|\,\yVec,\Theta)$ defined
by (\ref{EqnMarginalOne}) with (\ref{EqnSuffStat}). For simplicity, we use a Multi-Layer Perceptron (MLP) for the function mapping and trained with
cross-entropy. We use a generic MLP with one hidden layer. 
As such, the MLP itself is independent of the generative model considered but optimized for the generic truncated approximation (\ref{EqnMarginalOne})
which contains the model's joint. The idea of using a parametric function to approximate expectations w.r.t.\ intractable posteriors is an often applied technique \citep[e.g.][and refs therein]{GuEtAl2015,MnihRezende2016}.

\begin{wrapfigure}[21]{r}{0.60\textwidth}
\begin{minipage}{0.58\textwidth}
\ \\[-5mm]
\SetAlCapHSkip{0.2em}
\DecMargin{-1.0em}
\begin{algorithm}[H]\vspace{1.5mm}
%
initialize model parameters $\Theta$;\algBreak
for all $n$ init $\KKn$ such that $|\KKn|=S$;\algBreak
set $M_p$; (\# samples from prior distribution)\algBreak
set $M_q$; (\# samples from marginal distr.)\algBreak
\Repeat{$\Theta$ has sufficiently converged}{
\phantom{x=x}\ \\[-6mm]
update $M_p$ and $M_q$ (sampler adjustment)\algBreak
%
%
\For{($n=1,\ldots,N$)\vspace{1.5mm}}{
draw $M_p$ samples from $p(\sVec\,|\,\Theta) \rightarrow \KKn_p;$\algBreak
draw $M_q$ samples from $\qMarN(\sVec;\Theta)\rightarrow \KKn_q$,\hspace{10mm}
%
%
\algBreak
$\KKn\,=\,\KKn \cup \KKn_p \cup  \KKn_q$;\algBreak
remove those $(|\KKn|-S)$ elements\\
$\sVec\in\KKn$ with lowest $p(\sVec,\yVecN\,|\,\Theta)$;\vspace{2mm}\algBreak
}
$\KK=(\KK^{(1)},\ldots,\KK^{(N)})$;\vspace{1mm}\algBreak
use M-steps with (\ref{EqnSuffStat}) to change $\Theta$;\phantom{$\int_g$}\algBreak
}
\caption{Truncated Variational Sampling. \label{AlgTVS}
}
\end{algorithm}

\end{minipage}
\end{wrapfigure}
%

%
%
%
%

For our numerical experiments we combine prior and (approximate) marginal sampling to suggest new variational states. The easy to use prior samplers are not data driven and represent rather an {\em exploration} strategy. Marginal sampling, on the other hand, is rather an {\em exploitation} strategy that produces good results when sufficiently much from the data is already known. Mixing the two has therefore turned out best for our purposes.
Posterior samplers do require additional derivations but, to our experience so far, are also not necessarily better than combined prior and marginal sampling in optimizing the truncated free-energy.

Before we consider concrete generative models, let us summarize the general novel procedure in the form of the pseudo code given by Alg.\,\ref{AlgTVS}.
First, we have to initialize the model parameters $\Theta$ and the sets $\KKn$. While initializing $\Theta$ can be done as for other EM approaches, one 
option for an initialization of $\KKn$ would be the use of samples from the prior given $\Theta$ (more details are given below).
%
The inner loop (the variational E-step) of Alg.\,\ref{AlgTVS} is then based on a mix of prior and marginal samplers, and each of these samplers is directly defined in terms of a considered generative model, no model-specific derivations are used.
The same does not apply for the M-step but we will consider two examples how this point can be addressed: (A)~either by using well-known standard M-step or (B)~by applying automatic differentiation. Alg.\,\ref{AlgTVS} will be referred to as \emph{truncated variational sampling} (TVS).

\section{Applications of TVS}
Exemplarily, we consider two genrative models: Binary Sparse Coding and Sigmoid Belief Networks. The models are complementary
in many aspects, and thus serve well as example appications.

{\bf Binary Sparse Coding.} In the first example we will consider dictionary learning -- a typical application domain of variational EM approaches and sampling approaches in general. Probabilistic sparse coding models are not computationally tractable and common approximations such as maximum a-posteriori approximations can result in suboptimal solutions. 
Factored variational EM as well as sampling approaches have therefore been routinely applied to sparse coding. Of particular interest for our purposes are sparse coding models with discrete or semi-discrete latents \citep[e.g.][]{HaftEtAl2004,HennigesEtAl2010,GoodfellowEtAl2012,SheikhLucke2016}, where binary sparse coding \citep[BSC;][]{HaftEtAl2004,HennigesEtAl2010} represents an elementary example.
%
%
%
%
%
%
%
%
%
%
%
%

%
BSC assumes independent and identically distributed (iid) 
binary latent variables following a Bernoulli prior distribution, and it uses a Gaussian noise model:
\begin{eqnarray}
\disS{}p(\sVec\,|\,\Theta) \,=\, \prod_{h=1}^H\pi^{s_h}\,(1-\pi)^{1-s_h}\,,
%
\disS{}\phantom{iiiiii}p(\yVec\,|\,\sVec,\Theta) \,=\, \disS{\cal N}(\yVec;W\sVec,\sigma^2\One)\,,\label{EqnBSC}
%
%
%
\end{eqnarray}
where $\pi\in{}[0,1]$ and where $\Theta=(\pi,W,\sigma^2)$ is the set of model parameters.
%
%
%
%

As TVS is an approximate EM approach, let us first consider exact EM which seeks parameters $\Theta$ that optimize
the data likelihood for the BSC data model (\ref{EqnBSC}). Parameter update equations
are canonically derived and given by \citep[e.g.][]{HennigesEtAl2010}:
\begin{eqnarray}
&\disS\pi\,=\, \frac{1}{N}\sum_{n=1}^N\sum_{h=1}^H  \langle s_h \rangle_{q_n}, \label{EqnBSCMStepA}
\phantom{iiii}W\,=\, \Big(\sum_{n=1}^N \vec{y}^{(n)} \langle \vec{s} \rangle_{q_n}^{T}\Big) \Big(\sum_{n=1}^N \langle \vec{s} \vec{s}^{T} \rangle_{q_n}\Big)^{-1} \label{EqnBSCMStepB}\\
&\disS\sigma^2\,=\, \frac{1}{ND}\sum_{n=1}^N\langle \|\yVecN-W\sVec\|^2 \rangle_{q_n}\label{EqnBSCMStepC}
\end{eqnarray}
where the $q_n$ are equal to the exact posteriors for exact EM, $q_n=p(\sVec\,|\,\yVecN,\Theta)$.

A standard variational EM approach for BSC would now replace these posteriors by variational
distributions $q_n$. Applications of (mean-field) variational distributions as, e.g., applied by \citep[][]{HaftEtAl2004},
entails (A)~a choice which family of distributions to use; and (B)~additional derivations in order to derive update
equations for the introduced variational parameters. Also the derivation of sampling based approaches would require derivations. 
The same is not the case for the application of TVS (Alg.\,\ref{AlgTVS}). In order to obtain a TVS learning algorithm for BSC, we do (for the update equations)
just have to replace the expectation values in Eqns.\,\ref{EqnBSCMStepA} to \ref{EqnBSCMStepC} by (\ref{EqnSuffStat}). For the E-step, we then use the generative
model description (Eqns.\,\ref{EqnBSC})
in order to update the sets $\KKn$ using prior and approximate marginal distributions as described by Alg.\,\ref{AlgTVS}.
%
%
%
%
%
%
%
%
%
\begin{figure*}[h]
\begin{minipage}[c]{0.75\textwidth}
\centering
\includegraphics[width=0.99\textwidth]{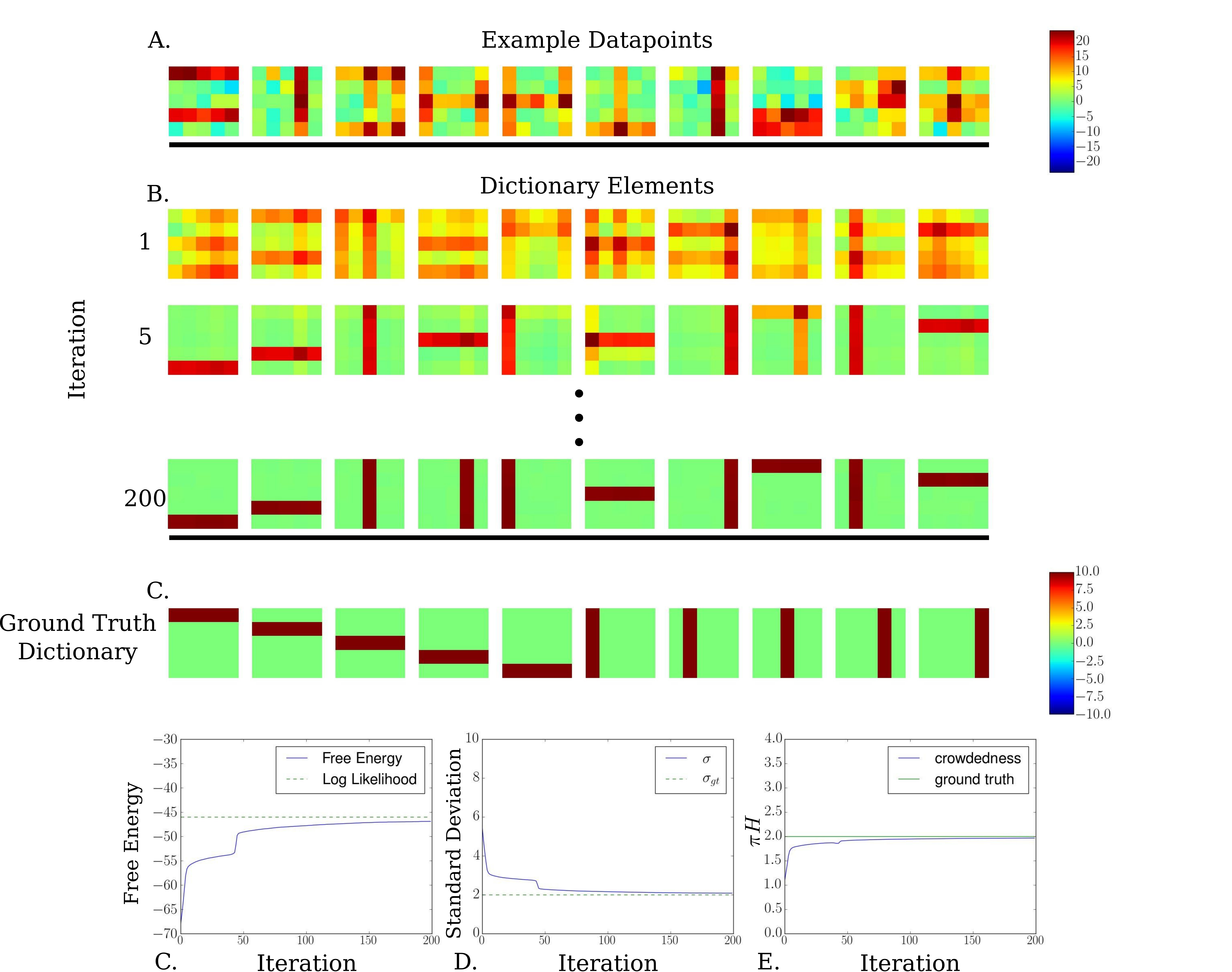}
\end{minipage}
\begin{minipage}[c]{0.22\textwidth}
\caption[Linear Bars Test]{\textbf{Linear Bars Test}. \textbf{A.} A subset
of the generated datapoints. \textbf{B.} The evolution of the dictionary
over TVS iterations. Note that permutations of the dictionary elements would yield
the same likelihood. \textbf{C.} The ground truth dictionary. \textbf{C.} The evolution 
of free-energy over TVS iterations plotted next to the exact log-likelihood. \textbf{D.} 
The evolution of the model standard deviation plotted next to the ground truth. \textbf{E.}
The evolution of the expected number of active units $\pi H$ plotted against the ground truth.}
\label{fig:BSCa}
\end{minipage}
\ \\[-1mm]
\end{figure*}

{\bf Artificial Data.} 
Firstly, we verify and study the novel approach
using artificial data generated by the BSC data model using ground-truth generating parameters $\Theta_{gt}$.
We use $H=10$ latent variables, $s_h$, sampled 
independently by a Bernoulli distribution 
parameterized by $\pi_{gt}=0.2$.
We set the ground truth parameters for the dictionary matrix,
$W\in \mathbb{R}^{D\times H}$ to appear like vertical and horizontal bars \citep[compare][]{HennigesEtAl2010} when rasterized
to $5\times 5$ images, see Figure \ref{fig:BSCa}, with a value of $10$ for a pixel that
belongs to the bar and $0$ for a pixel that belongs to the background. We linearly combine
the latent variables with the dictionary elements to generate a $D=25$-dimensional datapoint,
$\yVec$ to which we add mean-free Gaussian noise with standard deviation $\sigma_{gt}=2.0$. 
In this way we generate $N=10\,000$ datapoints that form our artificial dataset. 
We now use TVS for BSC to fit another instance of the BSC model to the generated data. 
The model is initialized with a noise parameter $\sigma$ equal to the average standard deviation 
of each observation in the data $\yVec^{(n)}$, the prior parameter is initialized as $\pi=1/H$ 
were the latent variable $H=10$ is maintained from the generating model. We initialize the columns 
of the dictionary matrix with the mean datapoint plus mean Gaussian samples with a standard 
deviation $\sigma/4$. 
We train the model using the TVS algorithm for $200$ TV-EM iterations maintaining the number of variational states
at $|\KKn|=S=64$ for all datapoints throughout the duration of the training. We use $M_{q}=32$ samples 
drawn from the marginal distribution (only 1st approximation) and $M_{p}=32$ samples drawn from the prior to vary $\KK$ according to Alg.\,\ref{AlgTVS}.
The evolution of the parameters during training is presented in Figure \ref{fig:BSCa}. We were able to extract
very precise estimates of the ground truth parameters of the dataset. Convergence is faster for the dictionary elements 
$W$ while we finally also achieve very good estimates for the noise scale $\sigma$ and prior $\pi$. 
We also appear to achieve a very close approximation of the exact log-likelihood using the 
truncated free-energy (Fig.\,\ref{fig:BSCa}), which shows that our free-energy bound is very tight for this data.

{\bf Image Patches.} For training, we now use $N=100\,000$ patches of size  $D=16\times 16$ from 
a subset of the Van Hateren image dataset \citep[][]{HaterenSchaaf1998} that excludes images
containing artificial structures. We used the same preprocessing as in \citep[][]{ExarchakisLucke2017}.
%
%
%
%
\begin{figure*}[t]
\begin{minipage}[c]{0.80\textwidth}
\centering
\includegraphics[width=0.99\textwidth]{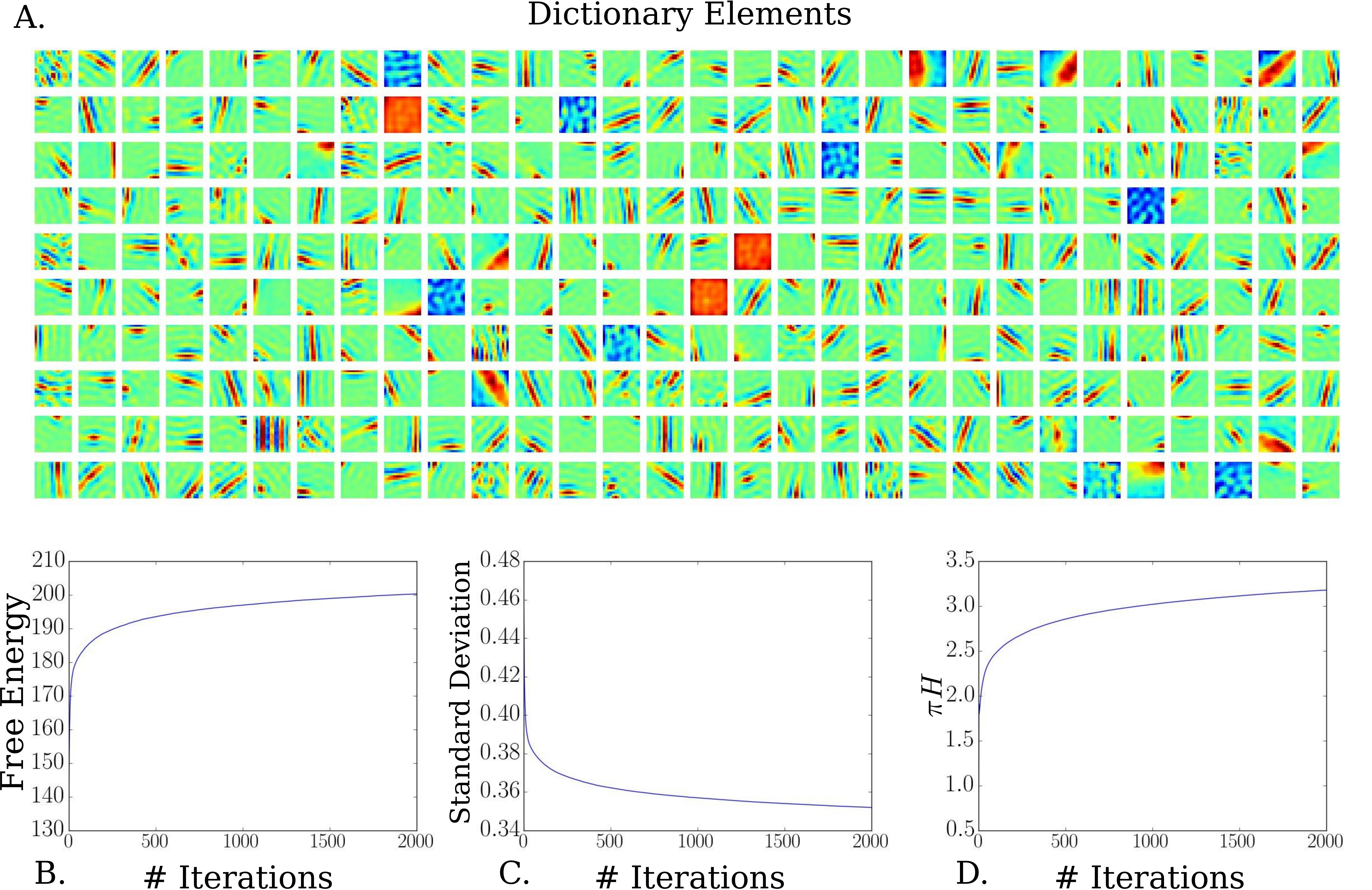}
\end{minipage}
\begin{minipage}[c]{0.19\textwidth}
\ \\[-17mm]
\caption[Natural Images]{\textbf{Image Patches}. \textbf{A.} The dictionary 
at convergence. \textbf{B.} The evolution of the free-energy over TVS iterations. 
\textbf{C.} The evolution of the model standard deviation over TVS iterations. \textbf{E.}
Evolution of the expected number of active units $\pi H$ over TVS iterations. Please enlarge for better visibility.}
\label{fig:BSCb}
\end{minipage}
\ \\[-2mm]
\end{figure*}
%
%
%
We trained BSC with TVS for $2000$ EM iterations and used a sampler adjustment (see Alg.\,\ref{AlgTVS}):
the first $100$ iterations used $M_p=200$ samples from the prior and $M_q=0$ samples from the
marginal distribution (only 1st approximation); from iteration $100$ to iteration $200$ we then linearly decreased the
number of prior samples to $M_p=0$ and increased the number of marginal samples to $M_q=200$ (at all times $M_p + M_q=200$).
%
Fig.\,\ref{fig:BSCb} shows the basis functions $W$ to converge to represent, e.g., Gabor functions \citep[compare][]{HennigesEtAl2010}.

%
%
%
%
%
%
%
%
%
%
%
{\bf Sigmoid Belief Networks.}
%
%
The second example we will consider here is a typical representative of a Bayesian Network: Sigmoid Believe Networks \citep[SBNs;][]{SaulEtAl1996}. While sparse coding approaches are applied to continuous (Gaussian distributed) observed variables, SBNs have binary observed and hidden variables. A further difference is that SBNs require gradients for parameter updates (partial M-steps), while parameter updates of sparse coding
models including BSC have well-known updates that fully maximize a corresponding free-energy (full M-steps). SBNs thus
serve as an example complementary to BSC, and is well suited for our purposes of studying generality and effectiveness of TVS. 

For simplicity, we will here consider an SBN with the same graphical model architecture as BSC: one observed and one hidden layer.
The SBN generative model is then given by:
\begin{equation}
\disS{}p(s_h) = \prod_{h} \pi_h^{s_h} (1-\pi_h)^{(1-s_h)},\phantom{iiiii}p(\vec{y}|\vec{s}) = \prod_{d} g_d ^{y_{d}} (1-g_d)^{(1-y_{d})} \label{EqnSBN}
\end{equation}
where $\pi_h$ parameterizes the prior distribution and where $g_d = \sigma(\sum_{h}W_{dh}s_h+b_{d})$ is a post-linear non-linearity with
Sigmoid function $\sigma$.

In general, inference for SBNs is challenging because of potentially show complex dependencies among its variables. Because of this, direct applications of
standard variational approaches \citep[e.g.][]{SaulEtAl1996} are challenging, and also popular recent variational methods applying reparameterization 
\citep[][]{KingmaWelling2014,RezendeEtAl2014} are not directly applicable. Also (variational) sampling approaches require additional mechanisms, e.g., the score function
based approach needs to reduce the variance of estimation \citep{MnihGregor2014}. 
%
%
%
%
%
%
\begin{figure}[h]
\ \\[0mm]
\centering
\begin{minipage}[c]{0.55\textwidth}
\centering
\ \phantom{iiiiiii}\includegraphics[width=0.99\linewidth]{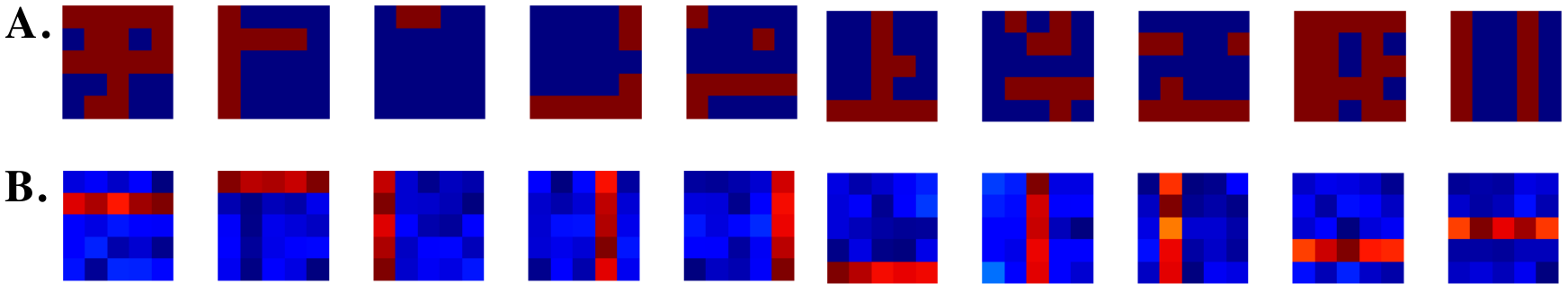}
\phantom{iiiiiii}\includegraphics[width=0.99\linewidth]{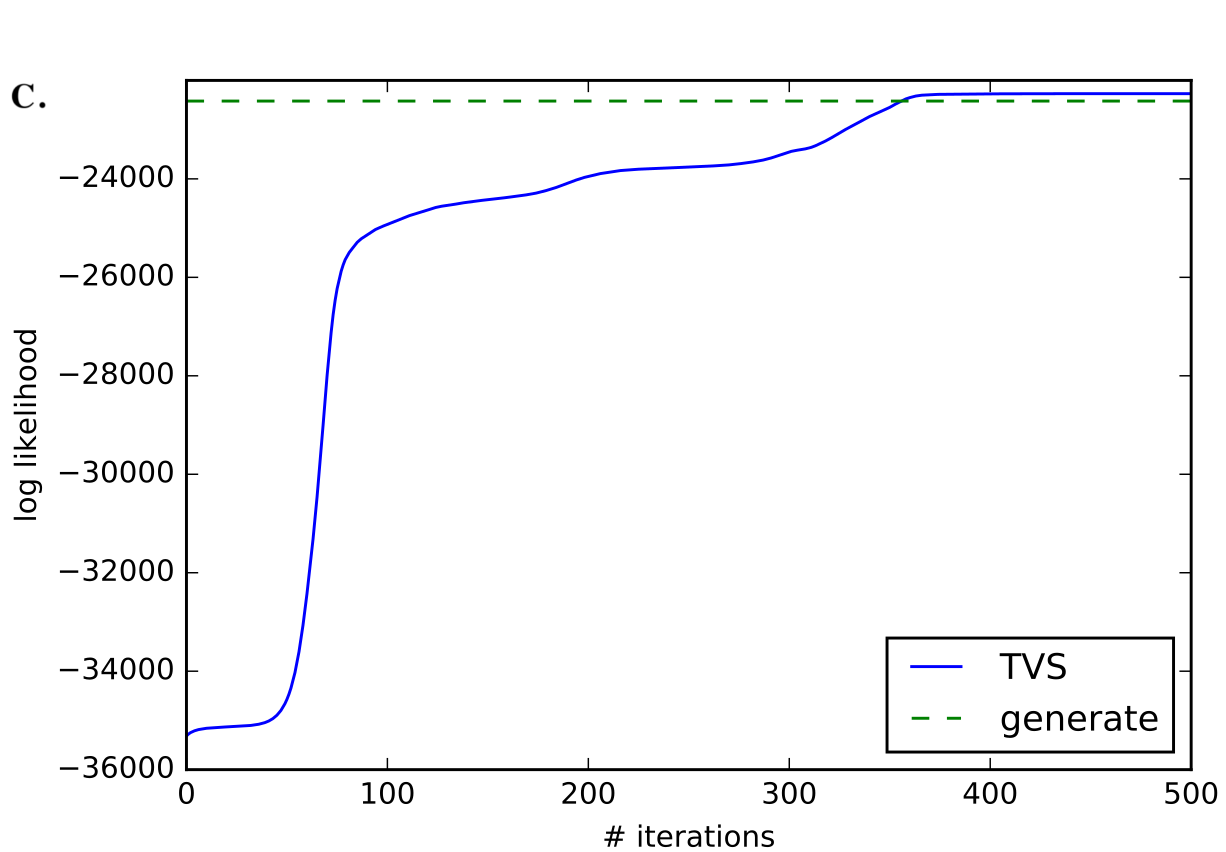}\vspace{-4mm}
\end{minipage}
\phantom{iiiiiiiii}
\begin{minipage}[c]{0.30\textwidth}
\caption{Application of a shallow SBN with TVS to artificial data (bars test). \textbf{A} Ten examples of the training data points. \textbf{B} Visualization of the learned weight matrix $W$ of the shallow SBN. All bars are discovered, one for each hidden unit. \textbf{C} Learning curve of the free energy. The dashed green line shows the true log likelihood of the SBN with the parameters used for generating the training data. Enlarge for better visibility.}
\label{fig:SBN_bars}
\end{minipage}
\ \\[-0mm]
\end{figure}

{\bf Artificial Data.} As for BSC, the optimization of SBNs by applying Alg.\,\ref{AlgTVS} does not require additional derivations.
We again first use a bars test as for BSC but the linearly superimposed bars now go through the sigmoid function and produce binary representations, i.e., generation according to (\ref{EqnSBN}).
We optimized an SBN with $H=10$ hidden units on $N=2000$ data points of such a bars test. For Alg.\,\ref{AlgTVS} we used $|\KKn|=S=50$ and very few samples for variation were found sufficient ($M_q=M_p=5$).
Results are shown in Fig.\,\ref{fig:SBN_bars}. The free energy (\ref{EqnFreeEnergy}) converges to even somewhat higher values than the (here still computable) ground-truth likelihood because of the limited size of training data. \vspace{-3mm}

%
%
\begin{table*}[t]
\begin{minipage}[c]{0.35\textwidth}
\centering
\ \\[-6mm]
\begin{tabular}{ccc}
\hline
    &       & approx. \\ 
model & \phantom{iii}$H$\phantom{iii}    & log-LL \\ \hline\hline
\phantom{$\int^f_b$}SBN (TVS)\phantom{$\int^a_b$}\hspace{-3mm}  & 100    & -121.91        \\
\phantom{$\int^a_b$}SBN (TVS)\phantom{$\int^a_b$}\hspace{-3mm}  & 200    & -111.23        \\\hline
\phantom{$\int^f_b$}SBN (Gibbs)$^*$\phantom{$\int^a_b$}\hspace{-3mm} & 200 & -94.3 \\
\phantom{$\int^a_b$}SBN (VB)$^*$\phantom{$\int^a_b$}\hspace{-3mm} & 200 & -117.0 \\
\phantom{$\int^a_b$}SBN (NVIL)$^{\diamond}$\phantom{$\int^a_b$}\hspace{-3mm} & 200 & -113.1 \\
%
SBN (WS)$^{\dagger}$\hspace{-3mm} & 200 & -120.7 \\
SBN (RWS)$^{\dagger}$\hspace{-3mm} & 200 & -103.1 \\
SBN (AIR)$^{\ddagger}$\hspace{-3mm} & 200 & -100.9 \\
\hline
\end{tabular}
\end{minipage}
\phantom{ii}
\begin{minipage}[c]{0.62\textwidth}
\ \\[-8mm]
\caption{Comparison of different models with two layers and different numbers of latents $H$ on binarized MNIST. ($*$) taken from \citep{GanEtAl2015}, ($\diamond$) from \citep{MnihGregor2014}, ($\dagger$) from \citep{BornscheinBengio2015}, ($\ddagger$) from \citep{Hjelm2016}. For TVS the final free energy on the test set can directly be computed by iterating the TV-E-step. The right column reports
these values as the estimated test log-likelihood (log-LL). Hence, for TVS the log-LL values are a lower bound estimation while results by \cite{BornscheinBengio2015,Hjelm2016} are from Monte Carlo estimations (and not necessarily lower bounds). For SBNs with 200 hidden units, we observed that TVS slightly outperforms NVIL, and its performance is comparable to RWS. The results of \citep[][]{GanEtAl2015} can not directly serveas a comparison of variational approaches (additional knowledge in the form of sparse priors were used).
\label{tab:sbn}\vspace{2mm}}
\end{minipage}
\ \\[-0mm]
\end{table*}
%

\ \\
{\bf Binarized MNIST.} Fianlly we apply SBNs to the Binarized MNIST dataset (downloaded from \citep{Larochelle2011}, converted as in \citep{MurraySalakhutdinov2008}). 
We used Alg.\,\ref{AlgTVS} with $|\KKn|=S=50$, $M_p=10$, $M_q=20$ (no sampler adjustment) and truncated marginal distributions approximated using an MLP with one hidden layer of $500$ hidden units and \emph{tanh} activation (2nd Approximation). Tab.\,\ref{tab:sbn} compares SBNs optimized by TVS with other models and optimization approaches.\vspace{-3mm}

\section{Conclusion\vspace{-0mm}}

The TVS approach studied here is different from previous approaches \citep[][]{RanganathEtAl2014,TranEtAl2015,RezendeMohamed2015,HoffmanBlei2015,KucukelbirEtAl2016}
as it does \emph{not} rely on a parametric form
of a variational distribution which is then, e.g., sampled from to estimate parameter updates. In contrast, for TVS, the drawn
samples \emph{themselves} define the variational distribution and act as its variational parameters. Changing the used
samples changes the variational distribution. TVS is thus by definition directly coupling sampling and variational EM which
in conjunction with its `black box' applicability is the main contribution of this study. 
One benefit of the tight
coupling seems to be that none of the diverse variance reduction techniques (which were central to BBVI or NVIL) are required.
TVS can thus be considered as the most directly applicable `black box' approach. We have here shown a proof of concept. More
advanced models and further algorithmic improvments will be the subject of future studies.\vspace{-2mm}
%
%
\bibliographystyle{splncs}

\end{document}